\newcommand{\edit}[1]{{\textcolor{black}{#1}}}
\documentclass{bmcart}

\usepackage{microtype}
\DisableLigatures[f]{encoding = *, family = *}
\usepackage{amsthm,amsmath}
\usepackage{color,soul}
\usepackage{multicol}
\usepackage{multirow}
\usepackage{graphicx}
\graphicspath{{figs/}}
\usepackage{textcomp}
\usepackage{xcolor}
\usepackage{booktabs}
\RequirePackage{natbib}
\RequirePackage{hyperref}
\usepackage[utf8]{inputenc} %unicode support
\startlocaldefs
\endlocaldefs

\begin{document}

%%% Start of article front matter
\begin{frontmatter}

\begin{fmbox}
\dochead{Research}


\title{Why I'm not Answering: Understanding Determinants of Classification of an Abstaining Classifier for Cancer Pathology Reports}

%%%%%%%%%%%%%%%%%%%%%%%%%%%%%%%%%%%%%%%%%%%%%%
%%                                          %%
%% Enter the authors here                   %%
%%                                          %%
%% Specify information, if available,       %%
%% in the form:                             %%
%%   <key>={<id1>,<id2>}                    %%
%%   <key>=                                 %%
%% Comment or delete the keys which are     %%
%% not used. Repeat \author command as much %%
%% as required.                             %%
%%                                          %%
%%%%%%%%%%%%%%%%%%%%%%%%%%%%%%%%%%%%%%%%%%%%%%

\author[
   addressref={LANL},                   % id's of addresses, e.g. {aff1,aff2}
   email={sayeradbl@lanl.gov}   % email address
]{\inits{SD}\fnm{Sayera} \snm{Dhaubhadel}}
\author[
   addressref={LANL},
   email={jamal@lanl.gov}
]{\inits{JMY}\fnm{Jamaludin} \snm{Mohd-Yusof}}
\author[
   addressref={LANL},
   email={kumkum@lanl.gov}
]{\inits{KG}\fnm{Kumkum} \snm{Ganguly}}
\author[
   addressref={LANL},
   email={gchennupati@lanl.gov}
]{\inits{GC}\fnm{Gopinath} \snm{Chennupati}}
\author[
   addressref={LANL},
   email={sunil@lanl.gov}
]{\inits{ST}\fnm{Sunil} \snm{Thulasidasan}}
\author[
   addressref={LANL},
   email={nickh@lanl.gov}
]{\inits{NH}\fnm{Nicolas} \snm{W. Hengartner}}
\author[
   addressref={LA},
%   email={@isuhsc.edu}
]{\inits{BJM}\fnm{Brent} \snm{J. Mumphrey}}         
\author[
   addressref={KY},
   email={ericd@kcr.uky.edu}
]{\inits{EBD}\fnm{Eric} \snm{B. Durbin}}
\author[
   addressref={UT},
%   email={}
]{\inits{EBD}\fnm{Jennifer} \snm{A. Doherty}}
\author[
   addressref={NJ},
%   email={}
]{\inits{EBD}\fnm{Mireille} \snm{Lemieux}}
\author[
   addressref={ORNL},
   email={schaefferknt@ornl.gov}
]{\inits{NS}\fnm{Noah} \snm{Schaefferkoetter}}
\author[
   addressref={ORNL},
   email={tourassig@ornl.gov}
]{\inits{GT}\fnm{Georgia} \snm{Tourassi}}
\author[
   addressref={IMS},
   email={CoyleL@imsweb.com}
]{\inits{LC}\fnm{Linda} \snm{Coyle}}
\author[
   addressref={DCCPS},
   email={lynnepenberthy.schumacher-penberthy@nih.gov}
]{\inits{LP}\fnm{Lynne} \snm{Penberthy}}
\author[
   addressref={LANL},
   email={mcmahon@lanl.gov}
]{\inits{BM}\fnm{Benjamin} \snm{H. McMahon}}
\author[
   addressref={LANL},
   corref={LANL},                       % id of corresponding address, if any
%   noteref={n1},                        % id's of article notes, if any
   email={tanmoy@lanl.gov}
]{\inits{TB}\fnm{Tanmoy} \snm{Bhattacharya}}
          
%%%%%%%%%%%%%%%%%%%%%%%%%%%%%%%%%%%%%%%%%%%%%%
%%                                          %%
%% Enter the authors' addresses here        %%
%%                                          %%
%% Repeat \address commands as much as      %%
%% required.                                %%
%%                                          %%
%%%%%%%%%%%%%%%%%%%%%%%%%%%%%%%%%%%%%%%%%%%%%%

\address[id=LANL]{%                           % unique id
  \orgname{Los Alamos National Laboratory}, % university, etc
%   \street{Waterloo Road},                     %
  \city{Los Alamos, NM},                              % city
  \postcode{87545}                                % post or zip code
  \cny{USA}                                    % country
}
\address[id=LA]{%
  \orgname{Louisiana Tumor Registry},
%   \street{D\"{u}sternbrooker Weg 20},
  \city{New Orleans, LA},
  \postcode{70122}
  \cny{USA}
}
\address[id=KY]{%
  \orgname{Kentucky Cancer Registry},
  \city{Lexington, KY},
  \postcode{40504}
  \cny{USA}
}
\address[id=UT]{%
  \orgname{Utah Cancer Registry},
  \city{Salt Lake City, UT},
  \postcode{84112}
  \cny{USA}
}
\address[id=NJ]{%
  \orgname{New Jersey State Cancer Registry},
  \city{New Brunswick, NJ},
  \postcode{08903}
  \cny{USA}
}
\address[id=ORNL]{%
  \orgname{Oak Ridge National Laboratory},
  \city{Oak Ridge, TN},
  \postcode{37831}
  \cny{USA}
}
\address[id=IMS]{%
  \orgname{Information Management Services Inc.},
  \city{Calverton, MD},
  \postcode{20705}
  \cny{USA}
}
\address[id=DCCPS]{%
  \orgname{National Cancer Institute, Division of Cancer Control and Population Sciences},
  \city{Bethesda, MD},
  \postcode{20850}
  \cny{USA}
}

%%%%%%%%%%%%%%%%%%%%%%%%%%%%%%%%%%%%%%%%%%%%%%
%%                                          %%
%% Enter short notes here                   %%
%%                                          %%
%% Short notes will be after addresses      %%
%% on first page.                           %%
%%                                          %%
%%%%%%%%%%%%%%%%%%%%%%%%%%%%%%%%%%%%%%%%%%%%%%

\begin{artnotes}
%\note{Sample of title note}     % note to the article
%\note[id=n1]{Equal contributor} % note, connected to author
\end{artnotes}

\end{fmbox}% comment this for two column layout

%%%%%%%%%%%%%%%%%%%%%%%%%%%%%%%%%%%%%%%%%%%%%%
%%                                          %%
%% The Abstract begins here                 %%
%%                                          %%
%% Please refer to the Instructions for     %%
%% authors on http://www.biomedcentral.com  %%
%% and include the section headings         %%
%% accordingly for your article type.       %%
%%                                          %%
%%%%%%%%%%%%%%%%%%%%%%%%%%%%%%%%%%%%%%%%%%%%%%

\begin{abstractbox}

\begin{abstract} % abstract
\parttitle{Background} %if any
Safe deployment of deep learning systems in critical real world applications requires models to make \edit{very} few mistakes, and only under predictable circumstances.  
%Development of such a model is \edit{a herculean task}.  
In this work, we address this problem \edit{using} an abstaining classifier \edit{that is} tuned to have $>$95\% accuracy, and \edit{then} identify the determinants of abstention using LIME (the Local Interpretable Model-agnostic Explanations method).  Essentially, we are training our model to learn the attributes of pathology reports that are likely to lead to incorrect classifications, albeit at the cost of reduced sensitivity.

\parttitle{Results} %if any
We demonstrate an abstaining classifier in a multitask setting \edit{for classifying} cancer pathology reports from the NCI SEER cancer registries on six tasks of interest.  For these tasks, we reduce the classification error rate by factors of 2--5 by abstaining on 25--45\% of the reports.  For the specific \edit{task} of \edit{classifying} cancer site, we are able to identify metastasis, reports involving lymph nodes, and discussion of multiple cancer sites as responsible for many of the classification mistakes, and \edit{observe} that the extent and types of mistakes vary systematically with cancer site ({\itshape e.g.,} breast, lung, and prostate).  When combining across three of the tasks, our model classifies 50\% of the reports with an accuracy greater than 95\% for three of the six tasks\edit, and greater than 85\% for all six tasks on the retained samples. Furthermore, we show that LIME provides a better determinant of classification than measures of word occurrence alone.

\parttitle{Conclusion} %if any
By combining a deep abstaining classifier with feature identification using LIME, we are able to identify concepts responsible for both correctness and abstention when classifying cancer sites from pathology reports. The improvement of LIME over keyword searches is statistically significant, presumably because words are assessed in context and have been identified as a local determinant of classification.
\\
\end{abstract}

%%%%%%%%%%%%%%%%%%%%%%%%%%%%%%%%%%%%%%%%%%%%%%
%%                                          %%
%% The keywords begin here                  %%
%%                                          %%
%% Put each keyword in separate \kwd{}.     %%
%%                                          %%
%%%%%%%%%%%%%%%%%%%%%%%%%%%%%%%%%%%%%%%%%%%%%%

% \begin{keyword}
% \kwd{sample}
% \kwd{article}
% \kwd{author}
% \end{keyword}

% MSC classifications codes, if any
%\begin{keyword}[class=AMS]
%\kwd[Primary ]{}
%\kwd{}
%\kwd[; secondary ]{}
%\end{keyword}

\end{abstractbox}
%
%\end{fmbox}% uncomment this for twcolumn layout

\end{frontmatter}

\section{Background}
\label{sec:intro}
Machine learning systems deployed for real world applications often encounter unforeseen situations that are not thoroughly explored during the model training. Such situations include data noise, variation in class composition, data quality, \edit{site-specific} and time-dependent definitions and processes, systematic and random label noise, and low-quality or inappropriate inclusion of data. The medical sector is a case in point, with numerous unknowns, and new concepts arising over time. It also exemplifies another typical constraint---a very high cost for mistaken classification. A sensible way to tackle these situations is to build a model that flags confusing or unusual data samples requiring human intervention, while classifying acceptable data samples \cite{hengartner2018}. While this process is intuitively obvious to humans, neural networks behave abnormally in many cases, making overconfident mistakes when encountering confusing or unknown inputs \cite{thulasidasan2019OnMT, nguyen2015deep}. 

In this work, we build a multitask abstaining classifier for classifying the text pathology reports from the US National Cancer Institute SEER (Surveillance, Epidemiology, and End Results) registries \cite{SEER}\edit. \edit{For a given report}, the classifier simultaneously makes predictions on six tasks of interest to the registries: primary site (70 classes), histological type (547 classes), primary  subsite (314  classes), laterality (7 classes), behavior (4 classes), and histological grade (9 classes). We use an extra class, called the abstention class, for each task and train the model to learn the features for each class including the abstention class. Unlike existing approaches for abstention that are typically based on softmax thresholding methods \cite{hendrycks2016baseline, bendale2016openmax}, our method learns features that lead to abstention \edit{\cite{thulasidasan2019}}, thus allowing us to understand the reasons causing the confusion. We use an interpretability technique, LIME \cite{ribeiro2016lime}, to understand the reasons \edit{for both}  predictions and abstentions made by our models.

We demonstrate how an abstaining classifier in a multitask setting can be used on real-world data to solve a complex problem and partially automate a human workflow, abstaining, {\it i.e.,} refusing to classify confusing samples, and \edit{hence} requiring human intervention on \edit{the abstained} samples while making predictions on regular samples without a problem. We show that this is an intuitively simple yet very effective way of tackling inevitable errors when deploying machine learning models. Using LIME, we demonstrate the trustworthiness and utility of our deep abstaining classifier (DAC) by visualizing the reasons for correct classifications as well as those for abstention.

\section{Results}
\label{sec:results}
Table~\ref{tab:abs_results1} shows the accuracy scores when two different models are trained on the \edit{data from} the Louisiana and Kentucky registries, first without and then with the `abstain' class but the same design otherwise. %The model is tested on the untouched\edit{/exclusive} test set from Louisiana and Kentucky registries and also on independent data from Utah and New Jersey registries. 
For each task of interest, \edit{we report the following}: the \edit{accuracy of the baseline classifier (when abstention is not applied) on all the test data}, the rate of abstention after applying abstention and tuning its $\alpha$ value \edit{(parameter that determines the penalty for abstention described in Section \ref{sec:abs_classifier}), the accuracy of the baseline classifier on the retained fraction, and the accuracy of the abstaining classifier} on the retained fraction of reports. \edit{These metrics are calculated for two different sets of data from different cancer registries}. These two sets, described in detail in Sections \ref{sec:dataset_description} and \ref{sec:exp_setup}, include four different cancer registries with one set having Louisiana and Kentucky registries (untouched test set different from train data)\edit, and another set having Utah and New Jersey registries. We \edit{observe} that the accuracy \edit{scores} and \edit{the} rates of abstention for Louisiana-Kentucky on the test set are consistent with the ones on the train data across all tasks. However, on \edit{the} Utah-New Jersey data, the accuracy scores are comparable but the abstention is slightly higher. This shows that the model generalizes well across different registries\edit, although at the cost slightly increased abstention when the input data looks different than what the model was trained on. The model achieves the desired accuracy of over \edit{95}\% for the individual tasks of predicting behavior, site, and laterality \edit{when abstention is constrained to be less than} 50\% \edit{on the holdout test data}. However, it fails to achieve the desired accuracy for the grade, histology, and subsite\edit, possibly because these are more complex problems with overlapping and changing definitions\edit, and have \edit{higher noise in the training data}. \edit{However the accuracy on the retained samples for the abstaining classifier is statistically identical to that of  the baseline classifer, suggesting that the main utility of the abstention class is to identify the \textit{confusing samples}}.

\begin{table}[htbp]
	\begin{center}
		\caption{Accuracy of baseline classifier with no abstention \edit{on the entire test set, and abstention rate, accuracy of baseline classifier} and accuracy of abstaining classifier on retained samples for individual tasks on data from four registries.} \label{tab:abs_results1}
		\begin{tabular}{l|c|c|c|c|c|c}
        \toprule
        & \multicolumn{4}{c|}{Louisiana - Kentucky} & \multicolumn{2}{c}{Utah - New Jersey} \\
        \hline
        \multirow{2}{*}\textbf{Task} & \textbf{Base Acc} & \textbf{\edit{Abst.}} & \textbf{\edit{Base Acc}} & \textbf{Accuracy} & \textbf{\edit{Abst.}} & \textbf{Accuracy} \\
                & \textbf{(no abs)} & \textbf{rate} & \textbf{\edit{(retained)}} & \textbf{(retained)} & \textbf{rate} & \textbf{(retained)} \\
        \midrule
        Behavior    &   97.91\%  &	 0.00\%  & \edit{97.91\%} &  97.85\%  &   0.00\%  &	96.63\% \\
        Grade       &   76.71\%  &	24.09\%  & \edit{83.21\%} &  83.35\%  &  29.70\%  &	78.20\% \\
        Histology   &   77.57\%  &	38.75\%  & \edit{90.23\%} &  90.27\%  &  47.36\%  &	87.88\% \\
        Laterality  &   91.34\%  &	43.94\%  & \edit{98.36\%} &  98.45\%  &  48.19\%  &	97.36\% \\
        Site        &   91.98\%  &	24.46\%  & \edit{98.75\%} &  98.80\%  &  28.90\%  &	98.05\% \\
        Subsite     &   65.11\%  &	20.41\%  & \edit{73.92\%} &  73.72\%  &  21.72\%  &	71.40\% \\
        \bottomrule
        \end{tabular}
	\end{center}
\end{table}

Likewise, Table~\ref{tab:abs_results2} shows the \edit{accuracy of the baseline model,} abstention rates, and the accuracy \edit{of the abstaining model} on the retained samples for different combinations of the tasks. This facilitates evaluation of the model based on the priority of the desired combination of tasks.
%for example, a registry may care more about getting site, behavior, and histology correct altogether but not care as much about getting site, behavior, and grade correct altogether. 
For any combination of tasks, a na\"\i{}ve guess for the accuracy and abstention rate would be the smallest of the individual accuracy and largest of the individual abstention rates, for example, for site-histology combination for LA-KY registries, the guess would be 90.27\% accuracy with 38.75\% abstention rate. However, this is possible only if the reports abstained by the model for the task with the \edit{largest} abstention rate is the \edit{super set} of the reports abstained by the model for the other tasks in the combination. Based on the results from Table~\ref{tab:abs_results2}, we see that the model abstains on different sets of reports for different tasks, making the combined accuracy lower and the combined abstention higher than the na\"\i{}ve guess.

\begin{figure}[htbp]
	\centering
%	\begin{minipage}{width=0.5\textwidth}
	\includegraphics[width=0.95\textwidth]{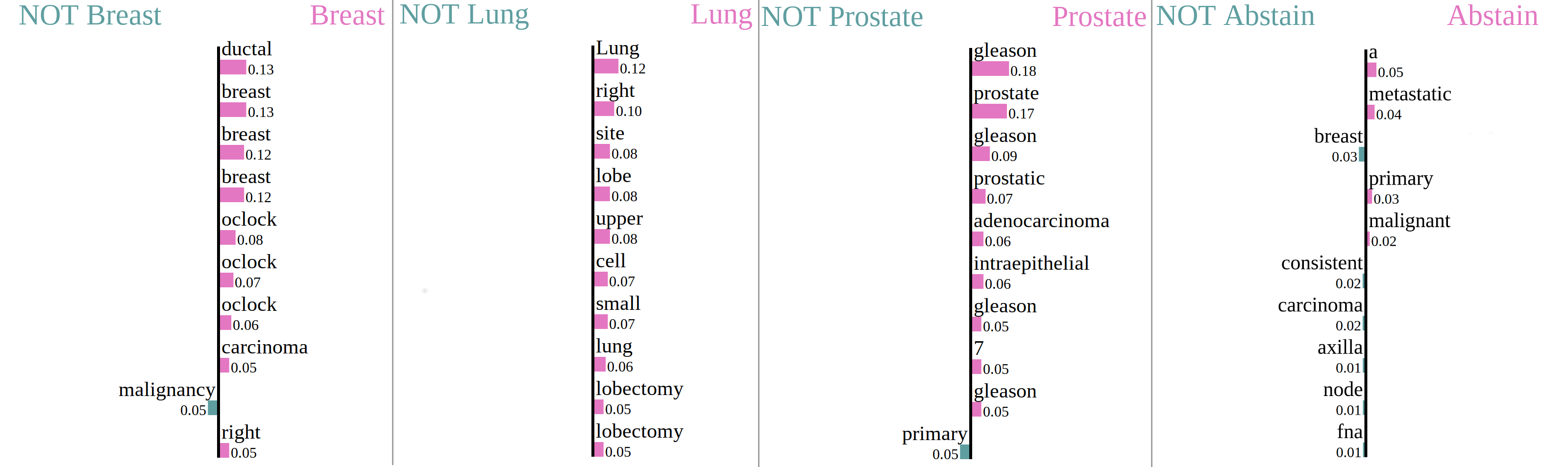}
	\caption{Typical LIME output for four pathology reports classified with the DAC. The top ten word instances \edit{(by magnitude of relative weight)} contributing to the indicated classification as assessed by LIME, together with the assigned coefficient. \edit{The coefficient values indicate the relative weight of the particular word instance in classifying the report as for or against a particular class.} Positive coefficients are indicated by magenta bars to the right of center, while negative coefficients are in blue-green, and they are ordered with most important words at the top.  Not shown, but provided by LIME is the context of each contributing word.}
	\label{fig:lime_out}
%	\end{minipage}
\end{figure}

\begin{table}[htbp]
	\begin{center}
		\caption{Accuracy of baseline classifier with no abstention \edit{on the entire test set,} and abstention rate and accuracy of abstaining classifier on retained samples for different combination of tasks on data from four registries. S:site, B: behavior, H: histology, L: laterality, G: grade}\label{tab:abs_results2}
		\setlength{\tabcolsep}{5pt}
		\begin{tabular}{l|c|c|c|c|c}
        \toprule
        & \multicolumn{3}{c|}{Louisiana-Kentucky} & \multicolumn{2}{c}{Utah-New Jersey} \\
        \hline
        \multirow{2}{*}\textbf{Task} & \textbf{Base Acc} & \textbf{\edit{Abst.}} & \textbf{Accuracy} & \textbf{\edit{Abst.}} & \textbf{Accuracy} \\
                & \textbf{(no abs)} & \textbf{rate} & \textbf{(retained)} & \textbf{rate} & \textbf{(retained)} \\
        \midrule
        S,B           &   76.71\%  &	24.46\%  &	96.61\%  &	28.90\%  &	94.56\% \\
        S,H           &   90.06\%  &	49.40\%  &	90.20\%  &	56.12\%  &	87.88\% \\
        S,B,H        &   71.74\%  &	49.39\%  &	89.51\%  &	56.12\%  &	86.41\% \\
        S,B,L        &   83.78\%  &	50.81\%  &	95.64\%  &	53.55\%  &	93.87\% \\
        S,B,G        &   70.09\%  &	40.77\%  &	82.46\%  &  47.68\%  &	76.78\% \\
        S,B,H,L    &   67.14\%  &	66.83\%  &	90.06\%  &	70.76\%  &	88.55\% \\
        S,B,H,L,G  &   53.19\%  &   72.24\%  &  76.06\%	 &  76.45\%  &	67.89\% \\
        \bottomrule
        \end{tabular}
	\end{center}
\end{table}
Figure~\ref{fig:lime_out} shows typical output from LIME for four specific reports.  We can see from the correctly classified cancer types that the results returned by LIME are logical: `prostate' is important when reports are correctly classified as prostate cancer, `lung' for lung cancer, and `breast' for breast cancer.  Additionally, LIME provides other words one might not have anticipated, but that make sense after examining the context.  For example, the subsite of breast tumors are often identified by analogy to a clock which is unique to breast cancers.  The top word for breast cancer (`ductal'), refers to a histology type (`ductal carcinoma') that is distinct to breast and pancreatic cancers.  Similarly, `lobes' and `lobectomies' are distinct to lung cancers, and `Gleason scores' to prostate cancer.  Also\edit, it is frequently the case that reports correctly assigned to a cancer type have mostly positive contributors to the classification.  

With the abstaining classifier, it is possible to identify those attributes of a report that suggest judgment should be withheld and the report be placed in the `abstain' class.  In the example shown, the word stem `metast\rlap', `lymph\rlap', and `node' are seen to be indicative of the abstention class, while words indicative of specific cancers weigh against the abstention class.  In the case shown, the abstained report was associated with a breast cancer, and we can see that LIME is weighing the decision to abstain vs. choosing the breast cancer class.  Also evident are several words that the DAC has evidently associated with abstention because of their association with metastasis: `metastatic' and `primary\rlap'. The ground truth available for training the DAC utilizes the annotations done by the SEER cancer registries, which assign reports to the original site of the cancer, so reports associated with metastasis are often difficult to correctly assign, and abstention is called for.

%https://www.overleaf.com/project/6050d93c9c4e595caf3be7fd

Our DAC was trained, validated, and tested on about 320,000 pathology reports.  This makes it difficult to assess general trends by hand.  By using both LIME and the abstention class, it is possible to understand more about how various concepts impact classification decisions.  In Table \ref{tab:LIME.words}, we provide statistical metrics to characterize the importance of eight distinct words in classifying three highly prevalent types of cancers, breast, lung, and prostate.  The words examined \edit{in the table were chosen from the list of 20 most common words picked up by LIME for each category such that they facilitate the distinction between correct and abstained reports from the perspective of a domain expert. For example, the most common LIME explanations for correctly classified breast cancer reports were -``breast", "ductal", ``left", ``o'clock", ``primary",``sentinel", ``metastatic", and ``carcinoma". Similarly, the most common LIME explanations for abstained breast cancer reports were: ``breast", ``metastatic", ``primary", ``carcinoma", ``right", ``lymphoma", ``node", and ``axillary". The keywords in the table} are suggested from the qualitative analysis described above, and include the cancer site (\textit{breast}, \textit{lung}, and \textit{prostate}), three words associated with metastasis (the word stem \textit{metast}, \textit{primary}, and \textit{origin}), two words associated with spread to lymph nodes (\textit{lymph} and \textit{node}), and two cancer-related words not specific to the site, (\textit{cancer} and \textit{malignant}).  For each of these words and cancer sites, we report in Table \ref{tab:LIME.words} the number of reports in which the word occurs, is ascribed \edit{importance} by LIME, and whether the most important occurrence for each word is for or against classification choice.  This is done for 320 correctly classified reports and 320 reports for which the classifier abstained on site classification, and is indicated, together with corresponding percentages in columns 3-8 of Table \ref{tab:LIME.words}.

\begin{table}[htbp]
    \centering
    \caption{Association of words with class choices of our DAC.  For the cancer type in column 1 and the word in column 2, we provide in columns 3 and 4 the number and percentage of reports with a given word when the report is correctly classified and when the report is assigned the abstention class.  In columns 5-6 and 7-8, we provide the number and percentage of the reports with the words where LIME picked up the word and when the coefficient for the word was positive for correctly classified and abstained reports. Column 9 is the p-value for word occurrence distinguishing abstaining and correct site class using 2x2 Fischer's exact test. Column 10 is the p-value for LIME identifying, and sign of LIME coefficient in distinguishing abstaining class from correct site identification with a 3x2 Fischer's exact test.  Columns 11 and 12 are p-values from binomial tests of whether the sign of the coefficients are associated with correctness and abstention.  Significance symbols used in Columns 9-12: ${\star}$ (p-value close to but below 0.05/24), ${\diamond}$ (p-value less than 0.01/24), and ${\ddagger}$ (p-value less than 0.001 divided by the number) indicate significance range.} \label{tab:LIME.words}
    \tiny
    \setlength{\tabcolsep}{4pt}
    %    \begin{tabular}{|p{3.1em}|p{3.0em}|p{2.7em}|p{2.3em}|p{2.8em}|p{2.6em}|p{2.7em}|p{2.2em}|p{3.8em}|p{4.1em}|p{3.8em}|p{3.8em}|}
    \begin{tabular}{|c|c|c|c|c|c|c|c|c|c|c|c|}
    \hline % headings                                                                          
    \multirow{3}{*}{\textbf{Site}}    & \multirow{3}{*}{\textbf{Word}} &\multicolumn{2}{|c|}{\textbf{Word in Report}}    &\multicolumn{4}{|c|}{\textbf{Word Highlighted by LIME} \textbf{\#(\%)}} & \multirow{2}{*}{\textbf{Occurr.}} & \textbf{LIME} & \textbf{LIME} & \textbf{LIME}\\
    \cline{3-8}
    \multirow{3}{*}{\textbf{\edit{(Corr./Abst.)}}} &  & \textbf{Corr.}  &  \textbf{Abst.}  &  \multicolumn{2}{|c|}{\textbf{Corr.}} & \multicolumn{2}{|c|}{\textbf{Abst.}}  &  \multirow{2}{*}{\textbf{p-value}}  &  \textbf{pickup}  &  \textbf{correct}  &  \textbf{abstain}\\
    \cline{3-8}
    % &&reports with word   &reports with word   &LIME ID      &\#(\% pos)        &\#(\%)     &\#(\% pos) \\
    && \textbf{\#(\%) }  &  \textbf{\#(\%)}  &  \textbf{ID}  &  \textbf{Positive}  &  \textbf{ID}  &  \textbf{Positive}  &  &  \textbf{p-value}  &  \textbf{p-value}  &  \textbf{p-value}\\
    % &&\#(\%)  &\#(\%)   &\#(\%)    &    &   & \\
    \hline
      & \textbf{breast} & 320(100) & 204(64) & 320(100) & 320(100) & 198(97) & 12(6) & 6.5e-41$^{\ddagger}$ & 4.3e-170$^{\ddagger}$  &   9.4e-97$^{\ddagger}$    &     2.9e-41$^{\ddagger}$\\
     & metast & 150(47) & 171(53) & 101(67) & 0(0) & 154(90) & 109(71) & 1.1-01 & 5.4e-39$^{\ddagger}$  &  7.9e-31$^{\ddagger}$   & 2.6e-07$^{\ddagger}$\\
     & primary & 103(32) & 117(37) & 97(94) & 0(0) & 109(93) & 87(79) & 2.8e-01 & 3.4e-37$^{\ddagger}$  &  1.3e-29$^{\ddagger}$   &  2.5e-10$^{\ddagger}$\\
    Breast & origin & 30(9) & 61(19) & 5(17) & 0(0) & 38(62) & 19(50) & 6.2e-4$^{\star}$ & 2.8e-08$^{\ddagger}$  &  6.2e-02   &    1.0  \\
    (320/320) & lymph & 230(72) & 177(55) & 8(3) & 6(75) & 115(65) & 36(31) &  1.8e-05$^{\ddagger}$ & 3.8e-31$^{\ddagger}$  & 2.95e-01    &     7.5e-05$^{\diamond}$ \\
     & node & 187(58) & 151(47) & 48(26) & 45(94) & 119(79) & 8(7) & 5.5e-03 & 2.2e-37$^{\ddagger}$  & 1.3e-10$^{\ddagger}$  &    2.5e-24$^{\ddagger}$\\
      & cancer & 172(54) & 59(18) & 28(16) & 2(7) & 36(61) & 19(53) & 7.6e-21$^{\ddagger}$ & 1.6-04$^{\diamond}$ & 3.0e-06$^{\ddagger}$ & 8.7e-01\\
     & malignan & 96(30) & 134(42) & 90(94) & 0(0) & 111(83) & 79(71) & 2.3e-03 & 3.7e-30$^{\ddagger}$  &  1.6e-27$^{\ddagger}$    &    9.4e-06$^{\ddagger}$\\
     \hline
    & \textbf{lung} & 296(93) & 146(46) & 296(100) & 296(100) & 139(95) & 44(32) & 1.8e-40$^{\ddagger}$ & 6e-104$^{\ddagger}$   &  1.6e-89$^{\ddagger}$     &    1.8e-05$^{\ddagger}$\\
     & metast & 127(40) & 208(65) & 91(72) & 7(8) & 176(85) & 55(31) & 1.9e-10$^{\ddagger}$ & 1.7e-15$^{\ddagger}$  &  7.1e-18$^{\ddagger}$    &   7.2e-07$^{\ddagger}$\\
     & primary & 111(35) & 99(31) & 95(86) & 35(37) & 75(76) & 29(39) & 3.5e-01 & 1.9e-01  &  1.3e-02      &   6.4e-02\\
    Lung & origin & 78(24) & 90(28) & 69(88) & 43(62) & 70(78) & 32(46) & 3.2e-01 & 1.4e-01  & 5.3e-02      &   5.5e-01 \\
    (320/320) & lymph & 148(46) & 110(34) & 29(20) & 13(45) & 68(62) & 21(31) & 2.8e-03 & 3.3e-05$^{\ddagger}$   &  7.1e-01    &    2.2e-03\\
     & node & 118(37) & 86(27) & 34(29) & 34(100) & 67(78) & 6(9) & 8.5e-03 & 7.8e-24$^{\ddagger}$  &  1.2e-10$^{\ddagger}$     &    1.5e-12$^{\ddagger}$\\
      & cancer & 59(18) & 56(18) & 29(49) & 28(97) & 26(46) & 3(12) & 8.4e-01 & 3.6e-11$^{\dagger}$  &  1.1e-07$^{\ddagger}$     &    8.8e-05$^{\diamond}$ \\
     & malignan & 135(42) & 194(61) & 94(70) & 12(13) & 109(56) & 38(35) & 4.2e-06$^{\ddagger}$ & 5e-04$^{\star}$  & 5.6e-14$^{\ddagger}$     &    2.0e-03 \\
     \hline
     & \textbf{prostate} & 319(100) & 239(75) & 318(100) & 317(100) & 210(88) & 20(10) & 1.2e-25$^{\ddagger}$ & 3e-153$^{\ddagger}$  & 1.2e-93$^{\ddagger}$    &     6.1e-36$^{\ddagger}$ \\
     & metast & 99(31) & 171(53) & 44(44) & 6(14) & 107(63) & 39(36) & 1.1e-08$^{\ddagger}$ & 3.5e-10$^{\ddagger}$  &  9.4e-07$^{\ddagger}$    &     6.5e-03\\
     & primary & 147(46) & 79(25) & 144(98) & 1(1) & 51(65) & 23(45) & 2.5e-08$^{\ddagger}$ & 7.6e-29$^{\ddagger}$  &  1.3e-41$^{\ddagger}$     &    5.7e-01\\
    Prostate & origin & 15(5) & 62(19) & 4(27) & 2(50) & 48(77) & 21(42) &  8.7e-09$^{\ddagger}$ & 4.3e-11$^{\ddagger}$  &  1.0e+00    &     4.7e-01\\
    (320/320) & lymph & 179(56) & 146(46) & 30(17) & 30(100) & 63(43) & 5(8) & 1.1e-02 & 1.3e-22$^{\ddagger}$  &  2.9e-01     &    7.5e-05$^{\diamond}$\\
     & node & 124(39) & 116(36) & 11(9) & 11(100) & 46(40) & 4(9) & 5.7e-01 & 3.4e-14$^{\dagger}$  &  9.8e-04     &    5.1e-09$^{\ddagger}$\\
      & cancer & 139(43) & 108(34) & 84(60) & 79(94) & 50(46) & 1(2) & 1.5e-02 & 6.8e-32$^{\ddagger}$  & 3.4e-18$^{\ddagger}$     &    9.1e-14$^{\ddagger}$ \\
     & malignan & 130(41) & 126(39) & 116(89) & 5(4) & 69(55) & 29(42) & 8.1e-01 & 2.0e-13$^{\ddagger}$  &   4.0e-27$^{\ddagger}$     &    2.3e-01\\
    \hline
    \end{tabular}
\end{table}

Not surprisingly, the cancer site is in all 320 breast and prostate cancer reports, and 93\% of the lung cancer reports, when the report is correctly classified.  The difficulty of relying on such key words in identifying the cancer site becomes evident, however, when one notices that between 46\% and 75\% of the abstained reports also contain these keywords.  This means that it is not simply lack of information, but other concepts, leading our DAC to abstain on reports.  LIME provides further insight.  In all except one report for prostate cancer, the cancer site is present in the report and counts in favor of the correct assignment, consistent with expectation from simple keyword searchers.  For abstained reports, we see that the most important instance of the words \textit{breast}, \textit{lung}, and \textit{prostate} appear in a context counting against abstention, although for lung cancer, one third of the instances of \textit{lung} appear in a context in favor of abstention.

In contrast, the word stem \textit{metast} is in less than half of the correctly classified, and more than half of the abstained, reports, consistent with it being an important source of confusion in understanding the primary site of cancer described.  With LIME, we see that, for the correctly identified reports, \textit{metast} is only important 40\% to 72\% of the time, and always counting against the correct classification.  For abstained reports, \textit{metast} is important 63\% to 90\% of the time, and counts in favor of abstention 31\% to 70\% of the time.  It is difficult to tell without further work whether LIME is successfully identifying those reports where the concept of metastasis is actually irrelevant to abstention.  We also see that the metastasis-related keywords \textit{primary} and \textit{origin} are less frequent, but convey similar information.

Discussion of \textit{lymph} and \textit{node} appear in approximately half of the reports, but are rarely identified as important by LIME, especially in breast and prostate cancers.  The word \textit{node} counts in favor of correct classification of lung and prostate cancers 100\% of the time, and 94\% of the time for breast cancers.  It nearly always counts against abstentions.  It may be that this is because the specific lymph nodes examined are characteristic of the cancer type.

The last four columns of Table \ref{tab:LIME.words} (columns 9-12) assign statistical significance to the association of keywords and LIME output to these assignments.  While occurrence is typically associated with correct classification or abstention by Fischer's 2x2 test, we see that for every word except one, the importance and sign of LIME identification has a stronger association, by Fischer's 3x2 test, with associations typically around $10^{-30}$ for LIME, compared to $10^{-3}$ for keyword occurrence.  Furthermore, with the last two columns of Table \ref{tab:LIME.words}, we present the statistical significance by which the LIME coefficient associates each word as in favor or against correct identification or abstention.

Examination of the counter-intuitive case in which \textit{lung}, \textit{breast}, or \textit{prostate} counts in favor of abstention shows that they are cases where multiple anatomical sites are discussed in the report.  It appears that our DAC is weighing the context of multiple sites being mentioned in the choice between abstention or classification; LIME appears to be able to identify which instances of word occurrences are motivating the DAC in its choice to guess or abstain.  

We observe that \textit{breast}, \textit{lung}, and \textit{prostate}, when considering the most important instance of each word, together with the context surrounding, are much more associated with correctly classified reports of their respective type and have positive weights when picked up by LIME compared to the abstained reports.  Column 9 of Table \ref{tab:LIME.words} provides the significance of association of each word with the abstention class.  Examination of the LIME results provides two more metrics of importance of keywords when they occur, as we can see from column 10 of Table \ref{tab:LIME.words}.  Cancer-specific terms are, not surprisingly, associated with reports correctly classified by site, rather than assigned to the abstention class.

\section{Discussion}
Modern machine learning techniques have revolutionized our ability to solve \edit{and automate} problems such as image detection and classification \cite{pham2020metapseudolabels, liu2021swin}, speech recognition \cite{zhang2020pushing}, language translation \cite{popel2020machinetranslation}, and natural language processing \cite{wolf2020transformers}. Typically, successful areas of application are data-rich and not amenable to mechanistic modeling or simple statistical techniques.  Another attribute of practical usage of machine learning techniques, however, is that they are used in error-tolerant environments.  There are a variety of reasons for this, with many of them reviewed by \cite{marcus2018criticaldeep}, and revolve around the difficulty of understanding the reasons the algorithm has used in making its prediction.  

In this work, we have examined the use case of cancer pathology report classification for the purpose of automating work flows for the SEER cancer registries.  Our results build a strong case for the utility of the seemingly obvious concept of abstention, that refrains \edit{from making} predictions on confusing samples while significantly improving the accuracy on the retained \edit{samples.} This is particularly useful in high\edit-consequence fields like medicine where expert intervention can be requested in case of doubt\edit{,} while automating a \edit{significant} fraction of the workflow. We demonstrate \edit{the ability to successfully identify confusing samples, although at the loss of some normal samples, and provide a} significant increase in accuracy across all the tasks \edit{on the retained samples. This is more likely to be acceptable for automation compared to the lower accuracy on all the samples.} Nevertheless, it is important to note that the desired accuracy is sometimes achievable only at a very high \edit{cost, causing} abstention on \edit{a} majority or even all of the samples. \edit{Likewise, even with a fixed value of the parameter $\alpha$ (described in Section \ref{sec:abs_classifier}) that determines the penalty for abstention, using the model on data from a different registry may lead to different abstention rates, as shown in Table \ref{tab:abs_results1} and \ref{tab:abs_results2}. Our previous work \cite{thulasidasan2019} shows that this is a property of \edit{the} DAC. A fixed $\alpha$ is not equivalent to a fixed abstention rate, but rather fixes the penalization coefficient that determines the abstention rate, which may be different for different sets of data depending on the level of noise and uncertainty present.} Moreover, in a multi-task setting like ours, getting multiple tasks correct simultaneously incorporates additional constraints and thus further reduces the coverage (\% retained). For instance, Table \ref{tab:abs_results2} shows that we can achieve 90\% accuracy on 33\% of the retained samples when requiring site, behavior, histology, and laterality to be correct simultaneously. On adding grade to this requirement, the accuracy drops to 76\% and \edit{achieves} only 28\% retention. We were able to reach the desired 97\% accuracy on 3 of the 6 tasks through a combination of a sophisticated DL model trained on a large (320K reports) data set and the use of an abstention category.  

\edit{Machine learning systems, while being able to achieve very high accuracy scores, are notorious for making mistakes on the most obvious predictions. Ribeiro et al. \cite{ribeiro2016lime} demonstrated how machine learning models are prone to picking up the artifacts in the data, questioning their trustworthiness, and highlighting the importance of understanding the reasons of prediction.  Sculley et al. \cite{sculley2015} also highlighted the long term burden machine learning models can cause, and the importance of a mechanistic understanding for predictions by discussing how the lack of strict abstraction rule to describe data, model, and predictions in machine learning causes huge technical over a long term due to instability in data, changing system configurations, entangling features making isolation of improvements impossible, and so on. Our use case of automating the classification pathology reports in a medical research database, that influences decisions in the medical sector as well as future research, requires the model to be both trustworthy as well as practical to maintain over a long time without a huge burden. Hence, to establish trust and have a mechanistic understanding of the model,} we were still left with the problem of understanding why the remaining errors were occurring and the circumstances leading to the abstention decision. The gold standard for assessing performance is hand curation, and indeed, we leveraged the hand-curated database to train our model.  Unfortunately, no such database exists for the question of \textit{why} our classifier was incorrect or chose to abstain, and we lacked sufficient resources to design and implement such a large-scale undertaking.  Furthermore, we wanted a system that could be transitioned into the cancer registries' workflow for providing evidence for classification decisions.

A variety of methods exist to provide evidence for \edit{the choices made by} DL classifiers, such as saliency maps \cite{simonyan2014deepinside, li2016visualizing, yuan2019}, SHAP \cite{lundberg2017shap}, and attention schemes \cite{choi2016retain}.  In general, these models have difficulty distinguishing complexity residing in the neural network layers from relationships distinct to a particular sample, as explained in \cite{adebayo2018saliencymapsproblem, serrano2019questioningattention, li2016visualizing}.  LIME addresses this problem by \edit{providing a separate explanation for each individual report in its own context. It does that} by systematically masking random sets of words from particular reports and examining which provide the greatest impact on a particular classification decision, possibly including the abstention class. While this technique was effective in providing the results shown here, it did require extensive sampling ($>$ 100,000 sample \edit{perturbations}) before results were stable enough to draw conclusions for a particular report. 

We chose to use LIME because text processing was an original use-case, our initial attempts with saliency maps were often difficult to interpret, perhaps for reasons described in \cite{adebayo2018saliencymapsproblem}, and it was readily adaptable to our DAC.  The original LIME use-case \cite{ribeiro2016lime} centered around movie reviews, and its evaluation involved much less complexity than our pathology report example.  We were able to develop a set of keywords that highlighted relevant concepts for both correctness and abstention, and show that LIME's ability to evaluate words in context provided more significant associations with both correctness and abstention than simple keyword searches. \edit{With the help of LIME explanations, we were able to manually observe three different kinds of noise in our data (text reports): no information, conflicting information, or incorrect information about the true label. Quantifying this, however, would also require significant hand curation so we deemed it to be out of the scope of this paper.}

\section{Conclusion}
We used a deep abstaining classifier to identify six attributes of cancers, by processing associated pathology reports.  By including an explicit abstention class \edit{and an appropriate associated loss function}, we were able to greatly increase the accuracy of classification on \edit{the} non-abstained reports. 

Regardless of whether or not we use abstention, it is always important to understand the reasons for a prediction and assess its credibility in real world. However, characterizing the reasons \edit{for} correctness and abstention in a statistically significant manner manually\edit{,} by reading individual pathology reports\edit{,} is highly labor intensive given the fact that we have about 320K reports. In addition, a lot of the reports are ambiguous, causing disagreements even among human expert reviewers regarding their correct classes. A simple keyword search of important class-related words for quantifying reasons of prediction, while sounds reasonable, does not take into account negations or contextual mentions. Through the use of LIME, we obtain the reasons of prediction of a significant fraction of reports automatically. The advantage of LIME over keyword search is that it takes into account the context of the words that are present in the reports and thus provides a basis for quantifying the reasons of correctness and abstention.

We showed, quite plausibly, through \edit{the use} of LIME, that reports were abstained when concepts such as metastasis or lymph nodes\edit{,} or multiple cancer sites were positively associated with the abstention class.  Identification of the determinants of abstention should facilitate our ability to use the DAC in a real-world setting. \edit{Hence, with the application of abstention and LIME, we were able to build a model that can be used partially automate a real-world workflow of classifying cancer pathology reports in the SEER registries. While this work is far from perfect in terms of achieving the ideal goal, we believe it definitely serves as a milestone in the application of machine learning in real world problems.}

\section{Methods}

\subsection{Dataset Description}
\label{sec:dataset_description}
The study was done on a corpus of text cancer pathology reports from the Louisiana and Kentucky Tumor Registries \edit{and also tested on reports from the Utah and New Jersey registries}. Each case of cancer (individual tumor), given by the \edit{case ID, was} identified by a combination of a \edit{patient ID and a tumor ID}\hbox{}. \edit{Ground truth was obtained for each case (tumor) of cancer from the manually abstracted and consolidated records in the cancer registries as there may be multiple reports for each tumor. Since the ground truth was consolidated for each individual tumor,} all the reports pertaining to a particular tumor \edit{had} the same ground truth regardless of the content of the text pathology report. \edit{The length of the reports were variable, but for the purpose of training the model, we use the commonly existing practice of making it a fixed length of 1500 word tokens by trimming the longer reports and appending zeros to the shorter ones.}

\subsection{Abstaining Classifier in a Multitask Setting}
\label{sec:abs_classifier}
A deep abstaining classifier \edit{\cite{thulasidasan2019}} or DAC, introduced first for combating label noise, is basically a regular deep neural network classifier (DNN) \cite{bengio2013DNN, schmidhuber2015DNN, lecun2015DNN} but with an extra (abstain) class and a custom loss function that permits abstention during training. This allows the DNN to \edit{identify and} abstain on (or decline to classify) confusing samples\edit{, without the need for manually labeling these cases,} while continuing to learn and improve classification performance on the non-abstained samples. 

\edit{For a given input $x$, denote $y$ to be the predicted class output by the DNN. We define $p_i = p_w(y=i|x)$ (the probability
of the $i\text{th}$ class given $x$) as the $i^\text{th}$ output of the DNN that
implements the probability model $p_w(y=i|x)$ (using a softmax function as its final layer ) with $w$ being the set of weight matrices of the DNN. For notational brevity, we use $p_i$ in
place of $p_w(y=i|x)$ when the input context $x$ is clear.}

\edit{The standard cross-entropy training loss for DNNs then takes the form
$\mathcal L_\text{standard} = -\sum_{i=1}^k t_i \log p_i$ where $t_i \in {0,1}$
is the target for the current sample.  The DAC has an additional $k+1^\text{st}$ output $p_{k+1}$ which is
meant to indicate the probability of abstention. We train
the DAC with the following modified version of the $k$-class
cross-entropy per-sample loss:
% The standard cross entropy loss function used in multi-class classification setting is given by, 
% \begin{equation}
%     \mathcal{L}_{standard} = - \sum_{i=1}^{k} t_{i}\log{p_{i}} \label{crossentropy_loss}
% \end{equation} where $t_i$ is the target for the current sample, and $p_i$ is the probability of i-th class. 
% The custom loss function for an abstaining classifier is a modified version of the standard cross-entropy and given by,
\begin{equation}
    \mathcal{L}(x) =(1-p_{k+1})(-\sum_{i=1}^{k}t_i\log\frac{p_i}{1-p_{k+1}}) + \alpha\log\frac{1}{1-p_{k+1}} \label{abs_loss}
\end{equation} where k is the number of classes excluding the abstention class, $t_i$ is the true label of training data for class i: it is one when $i$ is the true label, zero otherwise, $k+1$ is the abstention class,} $p_{k+1}$ is the probability of the abstention class and $\alpha$ is the penalty term for abstention.

This loss function behaves like a regular cross-entropy loss on the original classes and adds an additional loss \edit{term}, scaled by a tuning parameter $\alpha$ that controls the propensity \edit{for} abstention.   
% Since the balance between these terms depends on the data and the characteristics of the noise, the optimal value of $\alpha$ cannot be determined a priori and must be tuned during the training process. 
This parameter is tuned during training to guarantee an upper bound on the abstention rate while optimizing the accuracy. A very high value of $\alpha$ means \edit{a} high penalty for abstaining\edit{,} \edit{driving} the model towards no abstention\edit{. Conversely,} a very low value of $\alpha$ may drive the model to abstain on everything. %Ideally, we want to abstain on 0\% of the data and achieve 100\% classification accuracy 
\edit{The target abstention level is a function of the performance of the base classifier on the current dataset. For example, if the base classifier can obtain {\it e.g.,} 70\% accuracy on a given dataset, then the naive expectation might be that the DAC could then achieve perfect accuracy while abstaining on 30\% of the samples, but \edit{this} tends to be practically impossible because of the inherent noise in real world data.} 

At the start of \edit{training}, $\alpha$ is initialized to a \edit{small} value to encourage abstention on all but the easiest of examples learned so far. As the \edit{training} progresses on the true classes, \edit{t}he value for $\alpha$ is tuned 
%along with the model training 
with the goal of minimizing abstention and maximizing accuracy, with an upper bound on abstention rate. The trade-off between accuracy and abstention rates is explored by re-optimizing the network for varying abstention rates. \edit{It is important to note that $\alpha$ is not the same as the abstention rate but a penalty that determines the abstention rate for the data in hand; one can get different abstention rates for the same $\alpha$ value with different subsets of the data.}

%The DAC is tuned to simultaneously satisfy both a minimum accuracy on the non-abstained samples, as well as a maximum abstention rate which is derived from the performance of the base classifier, as best as it can. 
%If the base classifier can obtain {\it e.g.,} 70\% accuracy on a given dataset, then the naive expectation might be that the DAC could then achieve perfect accuracy while abstaining on 30\% of the samples. 
Thulasidasan et al.~\edit{\cite{thulasidasan2019}} reported that a DAC can learn unlabeled features in the data which may be correlated with label noise. In practice, the label noise is a mix of both uncorrelated ({\itshape e.g.,} labeling inconsistent with the report being classified) and correlated ({\itshape e.g.,} `metastasis' may indicate site labels are unreliable) so that perfect empirical identification of misclassified items can not be achieved.

We modify the multitask convolutional neural network (MTCNN) model by Alawad et al. \cite{alawad2019mtcnn} to include abstention for each task. Their model uses a word-level CNN \cite{krizhevsky2012cnn, chatfield2014cnn, kim2014cnn, szegedy2015cnn} in a multi-task learning setting \cite{collobert2008mtl, lee2010mtl, deng2013mtl, zhang2017mtl, zhang2018mtl} for automatic extraction of cancer information from unstructured text pathology reports to make predictions on 5 tasks: primary site (65 classes), laterality (4 classes), behavior (3 classes), histological type (63 classes), and histological grade (5 classes). Our model is an extension of their hard parameter sharing MTCNN where we train a model of similar architecture for six tasks (sub-site, in addition to the five previously listed) with a much higher number of classes per task. 

\begin{figure}[htbp]
	\centering
	  \includegraphics[width=0.95\textwidth]{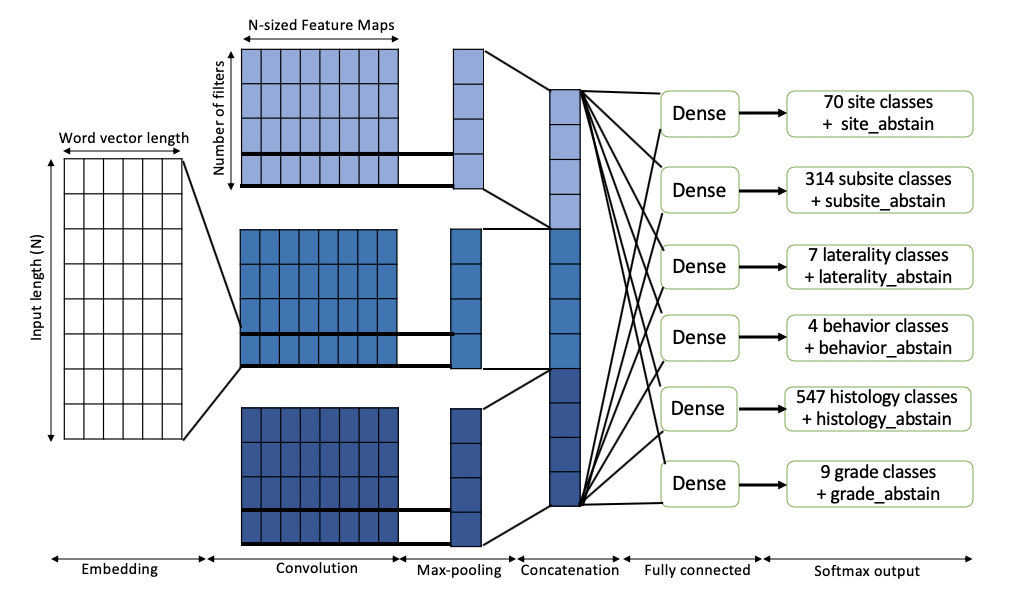}
	\caption{Architecture of our model; the model is largely similar to \cite{alawad2019mtcnn} other than an extra task and the additional abstain classes for each of the tasks.}
	\label{fig:model_architecture}
\end{figure}

The model diagram is shown in Figure~\ref{fig:model_architecture} where we can see \edit{an} extra `abstain' class for each of the tasks. The model has an embedding layer which represents each word token as a 300 dimensional word embedding \cite{mikolov2013embeddings, mikolov2013embeddingsICLR}. These vectors are fed to three independent convolutional layers followed by one-dimensional max pooling layers, each with 300 filters (or feature maps) and filter sizes of 3, 4, and 5 respectively. The outputs of these max pooling layers are then concatenated and fed to six independent fully connected layers with softmax output (one for each task) which return the predictions for each individual task.

\subsection{Experimental Setup}
\label{sec:exp_setup}
We train our abstaining classifier to achieve the following two goals that are desired by NCI and the cancer registries: i) a minimum of \edit{95}\% accuracy for each task on the retained samples,  ii) a maximum of 50\% abstention, where abstention on one task implies abstention on all tasks. In case the model cannot achieve both, we prioritize retention over accuracy and report the achieved accuracy on the retained samples. The model is trained on the pathology reports from Louisiana and Kentucky SEER registries (approximately 320K reports) with training-validation-test split of 60-20-20\%. After the training is done, using the validation set for setting the tuning parameters, we freeze all the parameters including $\alpha$ and evaluate the model on the test data and report the scores on these data alone. 
We report the model accuracy and abstention rates on the 20\% of the hold out test reports (untouched during training) from Louisiana and Kentucky registries. We further test the generalizability of the model across registries by reporting the accuracy and abstention of model prediction 
on pathology reports from Utah and New Jersey registries.

\subsection{Determinants of classification with LIME}
We use the Local Interpretable Model-agnostic Explanations (LIME) tool \cite{ribeiro2016lime} to identify which words (in context) were most important ({\em pro} or {\em con}) in determining the final class assigned to each pathology report.  LIME is provided with a trained DAC model and raw pathology reports.  We use the text version of LIME, requesting the top 40 words-in-context relevant to identifying the winning class, using 100,000 perturbations, as described in the LIME documentation.  These parameters resulted in a stable output, when comparing multiple runs on a sub-sample of the reports.

%%%%%%%%%%%%%%%%%%%%%%%%%%%%%%%%%%%%%%%%%%%%%%
%%                                          %%
%% Backmatter begins here                   %%
%%                                          %%
%%%%%%%%%%%%%%%%%%%%%%%%%%%%%%%%%%%%%%%%%%%%%%
\section*{List of Abbreviations}

SEER:
Surveillance, Epidemiology, and End Results
\\
NCI:
National Cancer Institute
\\
NLP:
Natural language processing
\\
DAC:
Deep Abstaining Classifier 
\\
LIME:
Local Interpretable Model-Agnostic Explanations
\\
CNN/CNNs:
Convolutional neural networks
\\
DL:
Deep learning
\\
LA:
Louisiana Tumor Registry
\\
KY:
Kentucky Cancer Registry
\\
UT:
Utah Cancer Registry
\\
NJ:
New Jersey State Cancer Registry

% \subsection{Authors' information (optional)}
% You may choose to use this section to include any relevant information about the author(s) that may aid the reader's interpretation of the
% article, and understand the standpoint of the author(s).

%%  After submission of the .TEX file,                     %%
%%  you will be prompted to submit your .BBL file.         %%

% if your bibliography is in bibtex format, use those commands:
\let\burl=\url
\bibliographystyle{bmc-mathphys} % Style BST file (bmc-mathphys, vancouver, spbasic).
\bibliography{bmc_article}      % Bibliography file (usually '*.bib' )
% for author-year bibliography (bmc-mathphys or spbasic)
% a) write to bib file (bmc-mathphys only)
% @settings{label, options="nameyear"}

% or include bibliography directly:
% \begin{thebibliography}
% \bibitem{b1}
% \end{thebibliography}

%%%%%%%%%%%%%%%%%%%%%%%%%%%%%%%%%%%
%%                               %%
%% Figures                       %%
%%                               %%
%% NB: this is for captions and  %%
%% Titles. All graphics must be  %%
%% submitted separately and NOT  %%
%% included in the Tex document  %%
%%                               %%
%%%%%%%%%%%%%%%%%%%%%%%%%%%%%%%%%%%

%%
%% Do not use \listoffigures as most will included as separate files

% \begin{backmatter}

\section*{Declarations:}

\subsection*{Ethics approval and consent to participate}
No ethics approval was required for the study.

\subsection*{Consent for publication}
Not applicable.

\subsection*{Availability of data and materials}
The data used in the analyses are health information legally protected against disclosure, and the property of the Louisiana, Kentucky, Utah, and New Jersey SEER cancer registries. Their use in the research has been approved by the appropriate authorities at the registries, the central DOE IRB, and the IRBs of the participating institutions.

\subsection*{Competing interests}
The authors declare that they have no competing interests.

\subsection*{Funding}
This work has been supported in part by the Joint Design of Advanced Computing Solutions for Cancer (JDACS4C) program established by the U.S. Department of Energy (DOE) and the National Cancer Institute (NCI) of the National Institutes of Health. This work was performed under the auspices of the U.S. Department of Energy by Argonne National Laboratory under Contract DE-AC02-06-CH11357, Lawrence Livermore National Laboratory under Contract DEAC52-07NA27344, Los Alamos National Laboratory under Contract DE-AC5206NA25396, and Oak Ridge National Laboratory under Contract DE-AC05-00OR22725. This research was supported by the Exascale Computing Project (17-SC-20-SC), a collaborative effort of the US Department of Energy Office of Science and the National Nuclear Security Administration.

\subsection*{Authors' contributions}
Sayera Dhaubhadel, Jamaludin Mohd-Yusof, and Benjamin H. Mcmahon did the bulk of the coding and writing of the manuscript. Brent J. Mumphrey, Eric B. Durbin, Jennifer A. Doherty, and Mireille Lemieux curated the data. \edit{Kumkum Ganguly annotated the data and helped with the writing.} Noah Schaefferkoetter and Georgia Tourassi made the baseline DL model available to this effort. Gopinath Chennupati, Sunil Thulasidasan, Nicolas Hengartner, and Noah Schaefferkoetter helped with the model validation and writing. Georgia Tourassi, Sayera Dhaubhadel, Linda Coyle, Lynne Penberthy, Benjamin H. Mcmahon, and Tanmoy Bhattacharya contributed to the study design and provided guidance and ideas throughout the project.

\subsection*{Acknowledgements}

We would like to acknowledge Elisabeth Moore for the initial suggestion to utilize LIME to provide interpretability of DL models for this effort.
% \end{backmatter}
\end{document}